\documentclass[11pt]{article}

\usepackage[preprint]{acl}

\usepackage{times}
\usepackage{latexsym}

\usepackage[T1]{fontenc}

\usepackage[utf8]{inputenc}

\usepackage{microtype}

\usepackage{inconsolata}

\usepackage{graphicx}

\usepackage{amsmath,amssymb}
\usepackage{booktabs}
\usepackage{multirow}
\usepackage{booktabs}
\usepackage{graphicx}
\usepackage{hyperref}
\usepackage{enumitem}
\usepackage{tikz}
\usepackage{subcaption}

\newcommand{\postspace}{\vskip -3mm}

\usetikzlibrary{shadows.blur}
\title{From Task Solving to Robust Real-World Adaptation in LLM Agents}

\author{Pouya Pezeshkpour\\
Megagon Labs\\
\texttt{pouya@megagon.ai} \\
\And
Estevam Hruschka\\
Megagon Labs \\
\texttt{estevam@megagon.ai}}

\definecolor{cadmiumgreen}{rgb}{0.0, 0.42, 0.24}
\definecolor{cardinal}{rgb}{0.77, 0.12, 0.23}
\definecolor{cadmiumred}{rgb}{0.89, 0.0, 0.13}

\usepackage[most]{tcolorbox}
\newtcolorbox[list inside=prompt,auto counter,number within=section]{prompt}[1][]{
    fontupper=\ttfamily\footnotesize,
    boxsep=5pt,
    left=0pt,
    right=0pt,
    top=0pt,
    bottom=0pt,
    boxrule=1pt,
    breakable,
    #1,
}

\begin{document}
\maketitle
\begin{abstract}
Large language models are increasingly deployed as specialized \emph{agents} that plan, call tools, and take actions over extended horizons. Yet many existing evaluations assume a “clean interface” where dynamics are specified and stable, tools and sensors are reliable, and success is captured by a single explicit objective—often overestimating real-world readiness. In practice, agents face underspecified rules, unreliable signals, shifting environments, and implicit, multi-stakeholder goals. The challenge is therefore not just \emph{solving} tasks, but \emph{adapting} while solving: deciding what to trust, what is wanted, when to verify, and when to fall back or escalate.
We stress-test deployment-relevant robustness under four operational circumstances: \emph{partial observability}, \emph{dynamic environments}, \emph{noisy signals}, and \emph{dynamic agent state}. We benchmark agentic LLMs in a grid-based game with a simple goal but long-horizon execution. 
Episodes violate clean-interface assumptions yet remain solvable, forcing agents to infer rules, pay for information, adapt to environmental and internal shifts, and act cautiously under noise.  
Across five state-of-the-art LLM agents, we find large gaps between nominal task-solving and deployment-like robustness. Performance generally degrades as grid size and horizon increase, but rankings are unstable: weaker models can beat stronger ones when strategy matches the uncertainty regime. Despite no explicit instruction, agents trade off completion, efficiency, and penalty avoidance, suggesting partial objective inference. Ablations and feature analyses reveal model-specific sensitivities and failure drivers, motivating work on verification, safe action selection, and objective inference under partial observability, noise, and non-stationarity.\footnote{\url{https://github.com/megagonlabs/wildgrid}}
\end{abstract}

\section{Introduction}

Large language models (LLMs) are rapidly evolving from passive task solvers into \emph{agentic} systems: models that plan, call tools, and take actions over extended horizons in complex environments \citep{comanici2025gemini,singh2025openai,luo2025large,dong2025survey}. As LLM capabilities have improved, a growing number of specialized agents have emerged, reporting strong performance on narrowly defined tasks \citep{zhang2024codeagent,novikov2025alphaevolve,hong2025data}. However, most evaluations still rely on two simplifying assumptions: (i) the environment is sufficiently specified and stable, and (ii) the objective is explicit and reducible to a clean success signal. In real-world deployment, these assumptions often break down, raising a central question: are today’s LLM agents truly ready for practical, in-the-wild use?

\begin{figure}
    \centering
    \includegraphics[width=\linewidth]{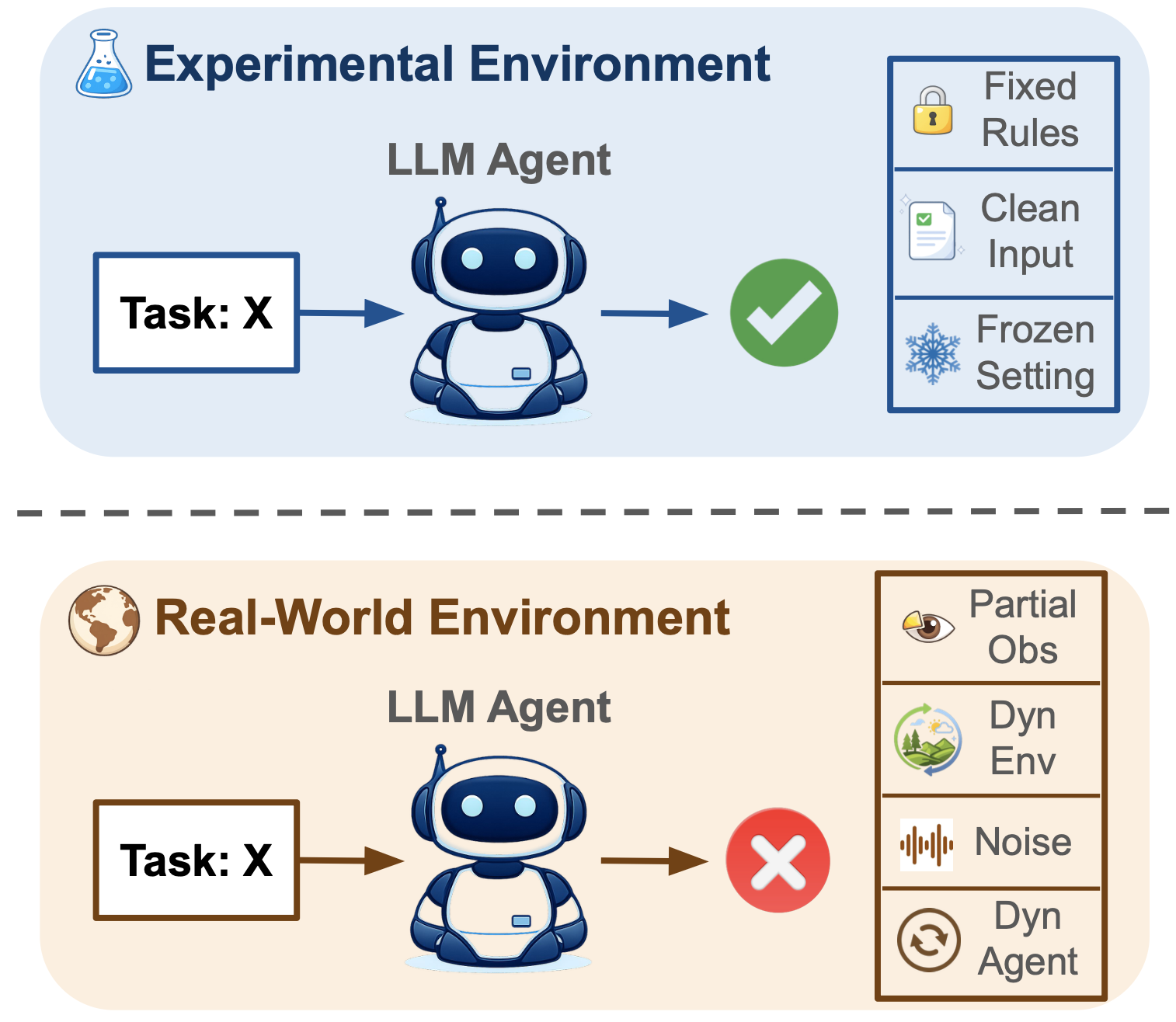}
    \caption{\textbf{Experimental vs. real-world deployment for LLM agents}. Agents often succeed in controlled experimental settings (fixed rules, clean inputs, frozen environment) but degrade in the real world due to partial observability, dynamic environments, noisy inputs, and changing agent behavior.
    }
    \label{fig:example}
    \postspace
\end{figure}

Our goal is not to ask whether an agent can solve a task in an idealized setting---specialized agents will soon achieve strong performance across many tasks. Instead, we ask: \emph{while solving the task, how robustly can an agent infer, adapt to, and safely interact with a realistic world?} We argue that real-world LLM agents will routinely face four circumstances that today’s benchmarks under-emphasize (Figure~\ref{fig:example}). First, \textbf{latent policies under partial observability and underspecified goals}: even a simple request (e.g., “bring supplies to hospital Room 312”) may require inferring unspoken constraints such as permissions, quiet-hour norms, and the trade-off between speed and disruption. Second, \textbf{dynamic environments and distribution shifts}: weather, detours, or new obstacles can invalidate prior plans and demand rapid replanning and safe fallback. Third, \textbf{noisy signals}: sensing and actuation can be unreliable (e.g., dropped frames, low traction, wear), requiring targeted verification when errors are costly. Fourth, \textbf{non-stationary agent state}: human overrides or model/tool updates can change behavior mid-task, forcing agents to detect drift and re-check assumptions.


To assess whether LLM agents are ready for real-world adoption, we introduce a grid-based puzzle with a simple objective but long-horizon execution, designed to intentionally violate clean-interface assumptions. The benchmark targets robustness during execution: inferring hidden rules, paying explicit costs to acquire information, adapting to environmental shifts and agent-state drift, and acting under noisy, partially observed inputs. The game is defined on an $N\times N$ grid with a small set of cell types and actions, but a clear goal: collect three key fragments ($K$) and exit through a door ($D$). At each step, the agent sees only a local window of the grid (centered on its current position), and these observations may be corrupted by noise. The grid includes rule tiles ($R$) whose behavior is undisclosed: interacting with $R$ triggers a latent transformation that may yield a key, a hazard, or nothing, depending on hidden context (e.g., current energy). This forces the agent to treat interaction as experimentation rather than executing a fixed plan.

The game operationalizes deployment stressors as controlled perturbations. The environment is non-stationary: hazards may spread, dynamics can change, and teleportation can occur mid-episode. 
The agent’s own capabilities can also drift (e.g., movement becomes less reliable or sensing becomes more expensive), requiring re-calibration. Finally, information acquisition is explicit and costly: \textsc{Scan} temporarily expands the local window of the grid visible by the agent, \textsc{Measure} collapses nearby latent cells ($\circ$) into known tiles, and \textsc{Interact} engages with the tile at the agent’s current location, triggering various outcome. Together, these mechanics form a compact testbed where success depends not just on reaching the exit, but on doing so robustly under partial observability, shifting dynamics, internal drift, and noisy signals. 

Evaluating five state-of-the-art LLMs, we observe a clear gap between nominal task-solving and deployment-like robustness. Performance typically degrades with larger grids and longer horizons, yet rankings are unstable: different models excel under different regimes, and in some settings a ``weaker'' model can outperform a stronger one when its strategy better matches the prevailing uncertainty. 
Beyond success rates, agents exhibit distinct trade-offs between completion, efficiency, and penalty avoidance, even though we never explicitly instructed them to do so, suggesting partial objective inference in some cases. Behavioral traces and single-stressor ablations reveal concrete failure drivers: some agents overcommit to early hypotheses about hidden mechanics, underinvest (or overinvest) in costly information gathering, and fail to recalibrate after shifts, while others adopt more verification-heavy strategies that improve efficiency but may reduce completion. Together, these results show that robustness in realistic, shifting environments depends as much on \emph{adaptive strategy selection} as on raw task-solving capability.

\section{Grid Game}
In this section, we introduce our interactive game setting, where the environment is underspecified and only partially observed, and where effective behavior requires inferring both what is true and what is wanted. We first describe the game’s core structure, then detail the modifiers that operationalize the four aforementioned real-world circumstances, and finally characterize the player (the LLM agent).

\subsection{Game Structure}
\begin{figure}[t]
\centering
\begin{tikzpicture}[x=0.6cm,y=0.6cm]

  \tikzset{
    celltxt/.style={font=\small\ttfamily},
    walltxt/.style={font=\small\ttfamily\bfseries},
  legendbox/.style={draw=brown!50, line width=0.45pt, rounded corners=3pt, fill=brown!8,
  blur shadow={shadow blur steps=5, shadow xshift=0.5pt, shadow yshift=-0.5pt}},
  legendtitle/.style={font=\small\sffamily\bfseries},
  legenditem/.style={font=\small\sffamily},
  }

  \draw[step=1, gray!60, very thin] (0,0) grid (9,9);
  \draw[black, line width=0.45pt] (0,0) rectangle (9,9);

  \foreach \x in {0,...,8} {
    \foreach \y in {0,...,8} {
      \node[celltxt, text=gray!70] at (\x+0.5,\y+0.5) {.};
    }
  }

  \foreach \x in {0,...,8} {
    \node[walltxt] at (\x+0.5,0.5) {\#};
    \node[walltxt] at (\x+0.5,8.5) {\#};
  }
  \foreach \y in {1,...,7} {
    \node[walltxt] at (0.5,\y+0.5) {\#};
    \node[walltxt] at (8.5,\y+0.5) {\#};
  }

  \node[celltxt] at (4.5,6.5) {h};
  \node[celltxt] at (5.5,6.5) {P};
  \node[celltxt] at (6.5,6.5) {R};

  \node[celltxt] at (2.5,5.5) {h};
  \node[celltxt] at (3.5,5.5) {?};

  \node[celltxt] at (2.5,4.5) {\#};
  \node[celltxt] at (3.5,4.5) {o};
  \node[celltxt] at (5.5,4.5) {\#};
  \node[celltxt] at (6.5,4.5) {h};

  \node[celltxt] at (3.5,2.5) {k};
  \node[celltxt] at (5.5,3.5) {h};

  \node[celltxt] at (4.5,2.5) {h};
  \node[celltxt] at (5.5,2.5) {?};
  \node[celltxt] at (6.5,2.5) {\#};

  \node[celltxt] at (1.5,7.5) {e};
  \node[celltxt] at (3.5,7.5) {h};
  \node[celltxt] at (7.5,7.5) {h};
  \node[celltxt] at (7.5,6.5) {D};
  \node[celltxt] at (7.5,4.5) {?};
  \node[celltxt] at (1.5,4.5) {R};
  \node[celltxt] at (1.5,2.5) {o};
  \node[celltxt] at (7.5,2.5) {k};
  \node[celltxt] at (6.5,1.5) {h};
  \node[celltxt] at (7.5,3.5) {\#};
  \node[celltxt] at (3.5,1.5) {\#};
    \node[celltxt] at (2.5,1.5) {h};
  \node[celltxt] at (5.5,7.5) {o};

  \fill[black!25, opacity=0.65, even odd rule] (0,0) rectangle (9,9)
                                                 (2,2) rectangle (7,7);

  \draw[black, line width=0.9pt] (2,2) rectangle (7,7);

  \fill[black]
    (4.5,4.72) -- (4.28,4.32) -- (4.72,4.32) -- cycle;

\draw[legendbox] (9.6,4.0) rectangle (12.6,9.0);
\fill[brown!25, rounded corners=3pt] (9.62,8.2) rectangle (12.58,8.98);
\node[legendtitle] at (11.1,8.6) {Actions};

\node[legenditem, anchor=west] at (9.6,7.8) {- Scan};
\node[legenditem, anchor=west] at (9.6,7.0) {- Measure};
\node[legenditem, anchor=west] at (9.6,6.2) {- Interact};
\node[legenditem, anchor=west] at (9.6,5.4) {- Move:};
\node[legenditem, anchor=west] at (9.9,4.6){E\textbackslash W\textbackslash S\textbackslash N};

\draw[legendbox] (9.6,0.0) rectangle (12.6,3.5);
\fill[brown!25, rounded corners=3pt] (9.62,2.7) rectangle (12.58,3.48);
\node[legendtitle] at (11.1,3.1) {State};

\node[legenditem, anchor=west] at (9.6,2.2) {- Step: 37};
\node[legenditem, anchor=west] at (9.6,1.45) {- Score: -6};
\node[legenditem, anchor=west] at (9.6,0.6) {- Energy: 4};

\end{tikzpicture}
\caption{A 9$\times$9 gridworld with partial observability: the agent observes only the centered 5$\times$5 window.}
\label{fig:game}
\end{figure}
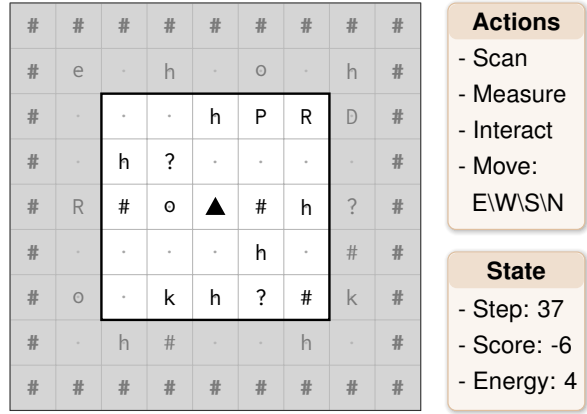

We define our episodic, turn-based grid game on a bounded \(N \times N\) world (see Figure \ref{fig:game}). Each episode begins by sampling a random seed and generating an initial map, which is fixed at initialization but may evolve over time due to dynamics introduced in later subsections. The agent occupies exactly one cell and maintains an \emph{orientation} (facing direction), rendered as \(\{\blacktriangle,\blacktriangledown,\blacktriangleleft,\blacktriangleright\}\). At every step, the agent receives a \emph{local} observation: a square window of radius \(r\) centered on its position (i.e., \((2r{+}1)\times(2r{+}1)\)). 
Episodes terminate when the agent successfully exits through the door, or when a fixed step budget (e.g., 200) is exhausted, in which case the outcome is a timeout/failure. The default objective is intentionally simple but long-horizon: the agent must collect \(K\) key fragments (by default \(K{=}3\)) and then open the exit door.

\paragraph{Tiles and placement.}
The grid uses a small vocabulary of discrete tile types. Border cells are always \emph{walls} (\(\#\)), forming an impassable boundary; interior layouts are generated by sampling additional walls and then carving corridors via short random walks to yield navigable maps. Remaining cells include \emph{empty} traversable space (\(\cdot\)), \emph{energy} tiles (\(e\)) that replenish the agent's energy when stepped on, \emph{key fragments} (\(k\)) that contribute to the win condition, a single \emph{door} (\(D\)) that ends the episode upon successful exit, \emph{hazards} (\(h\)) that incur penalties on contact, and \emph{rule tiles} (\(R\)) whose effects are undisclosed and triggered by interaction. The generator also places a small number of \emph{pads} (\(P\)) that act as stable landmarks and (when enabled) teleport landing points. Finally, some cells are \emph{Schr\"odinger tiles} we refer to as latent (\(\circ\)), representing unrevealed structure: they appear as \(\circ\) (or as ambiguous \((?)\) under observation noise) until the agent pays to resolve them (i.e., via \textsc{Measure}), after which they stochastically collapse into a standard tile (e.g., \(\cdot, e, R, k, h\)).
In our default instantiation, placement is controlled by simple density hyperparameters (e.g., sampling interior walls at a fixed rate followed by corridor carving, and populating small fractions of the remaining free cells with \(h\), \(e\), and \(R\)), together with fixed counts for unique objects (one door, one pad, one energy, two rules, and two key fragments). This design keeps the state representation compact and interpretable while making difficulty easily tunable by varying \(N\), densities, and the latent fraction.

\paragraph{Actions and effects.}
The agent's action space is small and discrete:
\(\textsc{Move}_{\{N,S,E,W\}}\), \(\textsc{Interact}\), \(\textsc{Scan}\), and \(\textsc{Measure}\) (when latent structure is enabled).
Movement updates both position and facing direction; moves can fail stochastically (``slip''), modeling actuation uncertainty. \(\textsc{Interact}\) applies to the single tile directly in front of the agent: interacting with the door attempts to open it (and succeeds only when the scenario requirement is satisfied, e.g., \(\ge K\) keys); interacting with a rule tile triggers a hidden transformation that may yield beneficial outcomes (e.g., \(k\)) or adverse outcomes (e.g., \(h\) or \(\cdot\)); and interacting with a hazard attempts to neutralize it at additional resource cost and with possible failure. \(\textsc{Scan}\) temporarily increases the observation radius for a one step duration, enabling broader situational awareness at an explicit energy cost. \textsc{Measure} collapses latent cells within a local radius, revealing hidden structure, and incurs an energy cost to reflect that probing can be expensive. Together, these tiles and actions define a compact but expressive environment where success requires long-horizon exploration, information acquisition, and risk-aware interaction---not just reaching a terminal goal under fixed, fully specified dynamics.

\subsection{Real-World Adaptations}

The Grid Game intentionally violates ``clean-interface'' assumptions common in agent evaluations. We operationalize four deployment-relevant challenges—latent policies under partial observability, dynamic environments, noisy signals, and dynamic internal agent state—via modular, parameterized mechanisms that are \emph{active during execution}, enabling controlled ablations and difficulty sweeps.

\paragraph{(1) Latent policies, partial observability and implicit objectives.}
Real deployments rarely expose complete state or fully specify what is permissible; agents must infer both \emph{what is true} and \emph{what is wanted} from limited and sometimes ambiguous evidence. We model this by restricting observations to a local window and introducing two sources of latent structure. First, \emph{rule tiles} (\(R\)) implement hidden, context-dependent transition dynamics: \textsc{Interact}ing with an \(R\) tile triggers an undisclosed transformation whose outcome depends on latent context---more specifically, if energy is high, it produces a key (with probability 0.5); if energy is low and a hazard is adjacent, it becomes a hazard; otherwise, it yields an empty tile. Second, \emph{latent cells} (\(\circ\)) represent unrevealed map content that can only be instantiated via an explicit probing action (\textsc{Measure}). Importantly, the game does not reveal the rule tables or collapse distribution; success therefore requires experimentation, hypothesis formation, and conservative decisions under ambiguity. Although the overall objective is simple, agents must also infer which interactions pay off long term and uncover two implicit goals: \textbf{minimizing steps} and \textbf{maximizing score}.

\paragraph{(2) Dynamic environment and distribution shift.}
To reflect the non-stationarity of real deployments, we introduce mid-episode perturbations that make the environment explicitly dynamic. At fixed intervals, the environment \emph{shifts}: a latent “weather” variable can change regimes, altering action reliability and resource drain, and periodic \emph{teleport} events can relocate the player to the pad tile. 
In addition, hazards evolve over time: at each step, hazard tiles can spread to a neighboring cell with some probability, creating local, compounding changes in the map. 
These interventions break the assumption of a fixed world model, forcing agents to detect regime changes, update beliefs, and adapt policies online.

\paragraph{(3) Noisy signals.}
Physical systems couple perception and action through noise, latency, and resource limits. We incorporate this by (i) corrupting observations with per-cell noise, producing occasional misread tiles and unknown symbols, and (ii) introducing stochastic action failures (e.g., movement ``slip''). We further model the cost of sensing and intervention through an explicit \emph{energy budget}: actions such as \textsc{Scan}, \textsc{Measure}, and \textsc{Interact} consume energy, and low energy amplifies actuation failure rates. 
Together, these constraints force agents to reason under uncertainty, choose verification strategically, and act risk-aware when signals are unreliable and mistakes are costly.

\paragraph{(4) Dynamic internal agent state.}
In deployed pipelines, the agent itself can change: models are updated, tools change behavior or cost, and internal calibrations drift. We simulate this by triggering a \emph{drift event} at a specified step during an episode. After drift, the same actions can have different reliability and cost profiles (e.g., higher movement slip probability, increased sensing costs), even if the external map appears unchanged. 
This forces agents to monitor outcomes and recalibrate assumptions about their own competence, rather than treating action models as fixed. Though simple, drift creates an ``I may have changed'' moment that punishes brittle plans and rewards adaptive, feedback-driven control.

\begin{table*}[t!]
\small
\centering
\begin{tabular}{lrrrrrrrrr}
\toprule 
\multirow{2}{*}{\bf Model} & \multicolumn{3}{c}{\bf 6$\times$6}&\multicolumn{3}{c}{\bf 8$\times$8}&\multicolumn{3}{c}{\bf 10$\times$10}\\
\cmidrule(lr){2-4}
\cmidrule(lr){5-7}
\cmidrule(lr){8-10}
&Acc&Score&Step&Acc&Score&Step&Acc&Score&Step\\
\midrule
GPT-5.2&48.0&5.5&35.1&40.0&- 4.4&51.1&26.0&- 28.3&85.2\\
GPT-5 mini&34.0&\bf6.4&\bf23.2&22.0&- 0.5&\bf 31.9&30.0&\bf - 2.2&\bf 41.7\\
Gemini-3 Pro&\bf 50.0&-0.2&43.12&28.0 & \bf 5.0 & 37.4&\bf 38.0 & -22.3 & 73.4\\
Gemini-3 Flash&48.0 & 3.7 & 30.8&\bf 42.0 & -3.1 & 54.6 & 32.0 & -8.8 & 52.7\\
Qwen3&2.0&- 98.0&110.0&0.0&-&-&0.0&-&-\\
\bottomrule
\end{tabular}
\caption{Main results with all modifiers active. We report success rate (\textbf{Acc}), average \textbf{Score}, and \textbf{Steps} per grid size (Score and Steps are averaged over successful episodes). Higher Acc/Score and lower Steps are better; Steps and Score capture implicit efficiency and reward objectives not explicitly optimized in the prompt.}
\label{tab:main_res}
\postspace
\end{table*}

\subsection{The Player}

The \emph{player} is an LLM-based agent that interacts with the Grid Game through a text-only interface. At each timestep, the environment returns a structured state summary rendered as a compact textual observation, without privileged access to the full map, hidden rules, or latent variables. Specifically, we provide: (i) the overall \emph{goal}, (ii) a \emph{local view} centered on the agent with its facing direction, (iii) a \emph{state vector} (step, remaining energy, keys collected, score), and (iv) the \emph{available actions} (i.e., \(\textsc{Move}_{N,S,E,W}\), \textsc{Interact}, \textsc{Scan}, and \textsc{Measure}). In addition, we pass a short history of the agent's previous actions to facilitate long-horizon reasoning and enable models to recognize non-stationarity without requiring external memory beyond the prompt.

We also provide a minimal \emph{execution log} that records salient events and outcomes. At each timestep, the environment appends a log entry (when something noteworthy occurs) that includes: (a) step metadata (timestep index, chosen action, and a local $3\times3$ view), (b) whether the action succeeded (e.g., ``MOVE\_E succeeded'' vs.\ ``MOVE failed (slip)''), (c) immediate rewards or penalties (e.g., energy gained from $e$, penalties from hazards), and (d) notable environment events (e.g., ``ENV SHIFT,''). 
This event-based logging serves two purposes: it mirrors real deployments where agents must interpret telemetry and tool responses rather than observe latent state directly, and it makes both positive and negative outcomes explicit without reducing the task to a single scalar reward. 
We instantiate the LLM player with a standardized prompt enforcing an action-only output format; the full prompt is provided in the Appendix.

\section{Experimental Details}
We evaluate five state-of-the-art LLMs: \textsc{GPT-5.2}, \textsc{GPT-5 mini} \citep{singh2025openai}, \textsc{Gemini 3 Pro}, \textsc{Gemini 3 Flash} \citep{comanici2025gemini}, and \textsc{
Qwen3-235B-A22B} \citep{yang2025qwen3} (non-thinking). For models that support explicit test-time ``thinking,'' we use their default thinking budget (medium or high). 
For the main evaluation, we generate 50 random game instances for each grid size in
$\{6\times6,8\times8,10\times10\}$. Except for grid size, we use the same parameters across samples from different grid sizes to ensure a fair comparison. 
In each instance, we sample key environment parameters uniformly at random: We sample $\texttt{observation noise}\!\sim\!\mathcal{U}(0,0.2)$, $\texttt{move fail}\!\sim\!\mathcal{U}(0,0.1)$, and $\texttt{latent fraction}\!\sim\!\mathcal{U}(0,0.2)$. 
We fix the remaining dynamics across all main experiments: a $5\times5$ observation window centered on the player, environment mode shifts every 25 steps, teleport events every 50 steps, and agent-state drift every 100 steps.  
For ablations, we generate 5 instances for each data point per condition. In each ablation, we deactivate all perturbations and vary only the single factor under study, to isolate its causal impact on performance. 

\section{Experiments}
We begin by evaluating LLM agents in the full game setting with all modifiers active, reporting average performance over 50 randomly generated episodes per grid size. We then analyze strategic behavior by comparing how models allocate actions over time during an episode. Next, we run single-stressor ablations, varying one modifier at a time while disabling all others to quantify its isolated impact on performance. Finally, we pool episodes across all experiments and fit per-model logistic regression predictors over a 9-dimensional feature vector to attribute win/loss outcomes to specific environment and dynamics factors.

\begin{figure*}[th]
    \centering
    \begin{subfigure}[b]{0.32\textwidth}
        \centering
        \includegraphics[width=\linewidth]{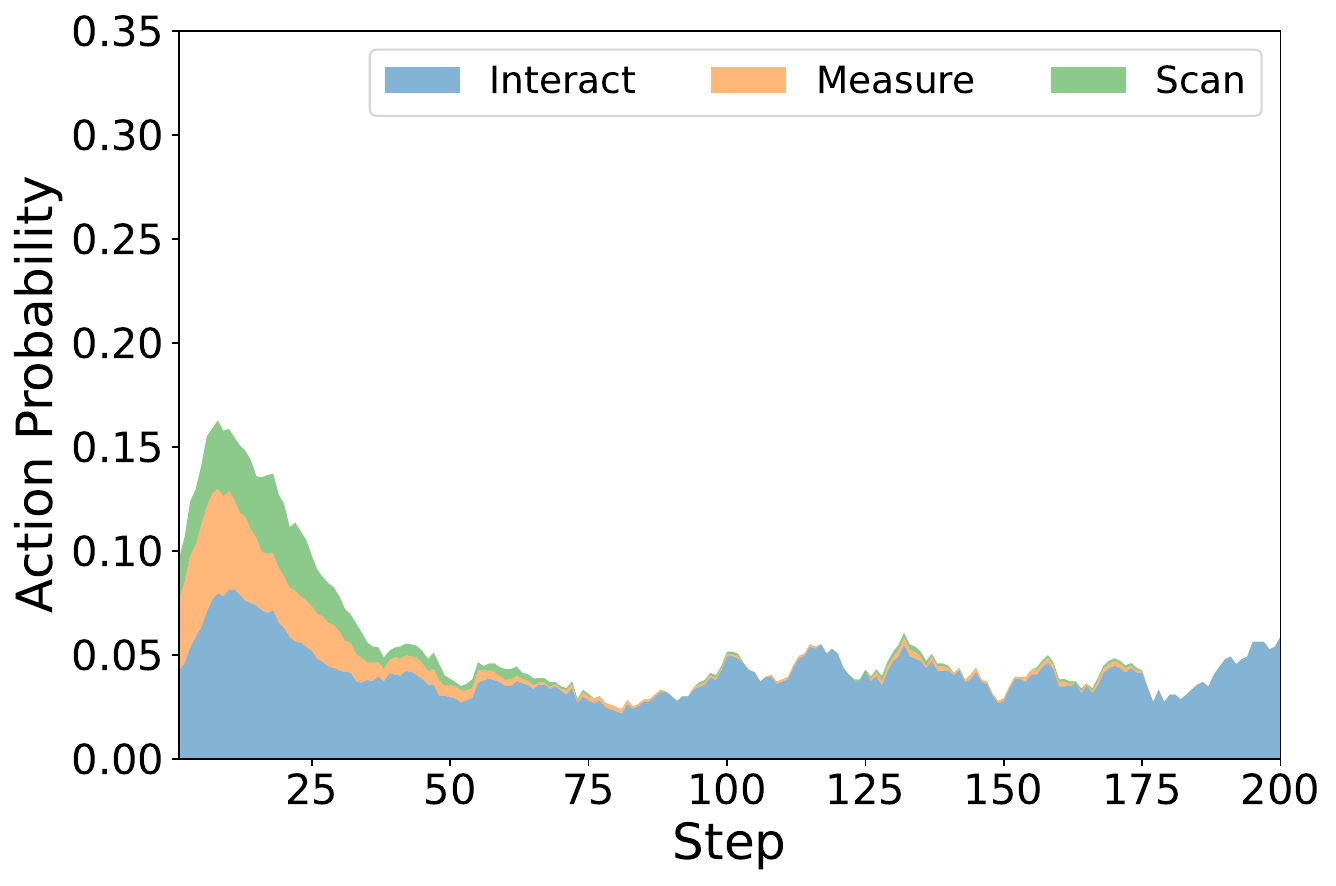}  
        \caption{GPT-5.2}
    \end{subfigure}
    \begin{subfigure}[b]{0.32\textwidth}
        \centering
        \includegraphics[width=\linewidth]{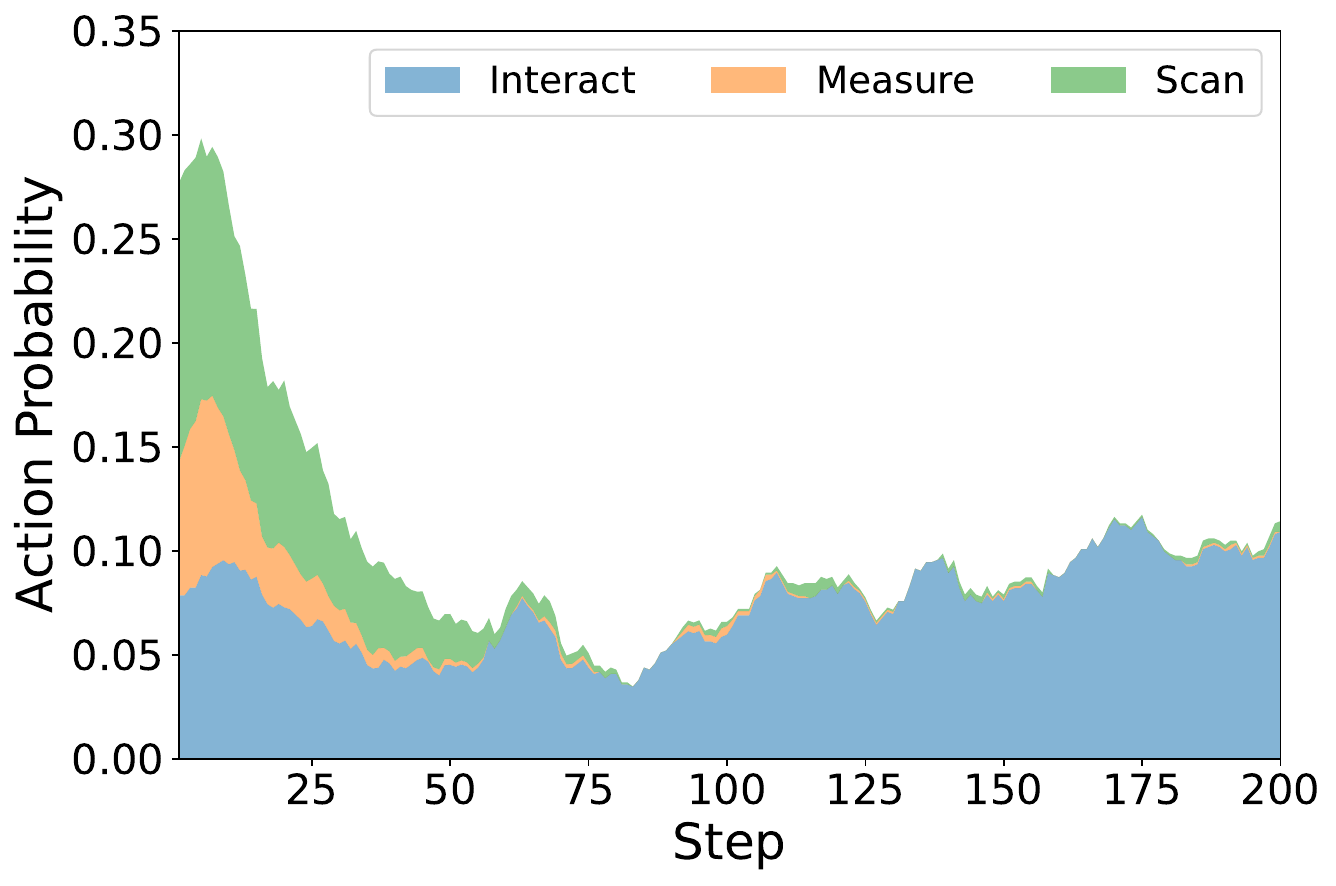}  
        \caption{GPT-5-mini}
    \end{subfigure}
    \begin{subfigure}[b]{0.32\textwidth}
        \centering
        \includegraphics[width=\linewidth]{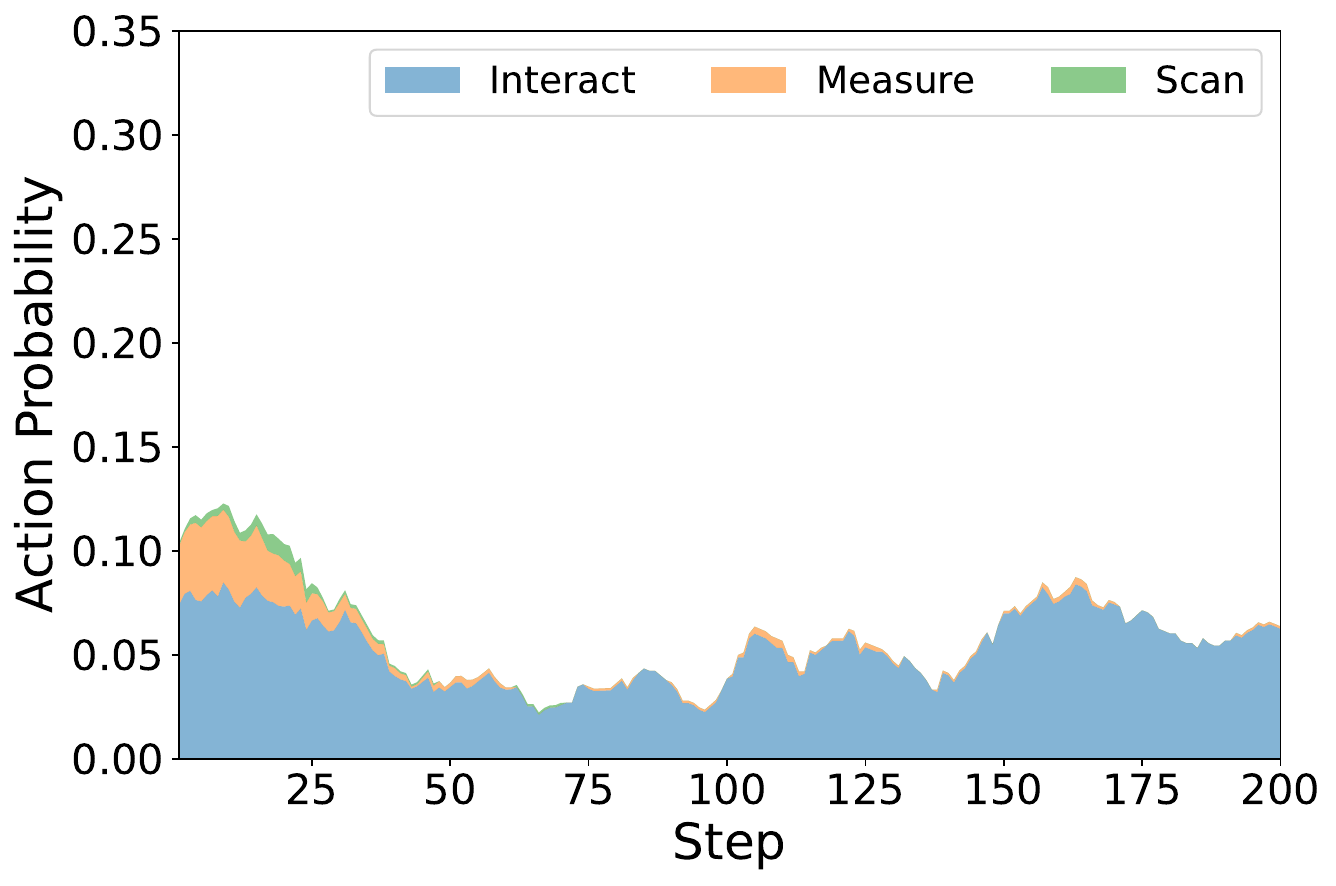}
        \caption{Gemini-3 Pro}
    \end{subfigure} 
    \begin{subfigure}[b]{0.32\textwidth}
        \centering
        \includegraphics[width=\linewidth]{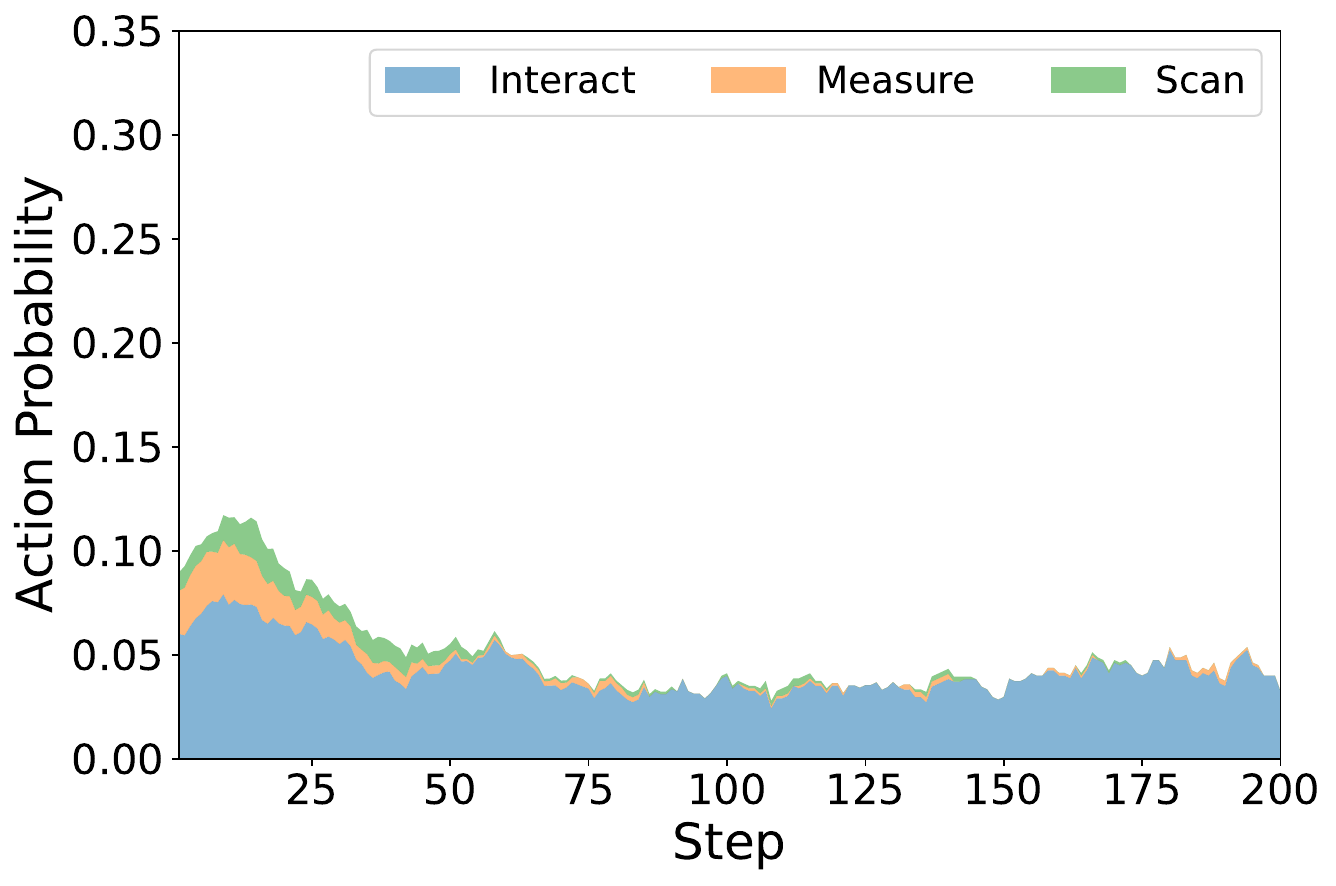}  
        \caption{Gemini-3 Flash}
    \end{subfigure}
    \begin{subfigure}[b]{0.32\textwidth}
        \centering
        \includegraphics[width=\linewidth]{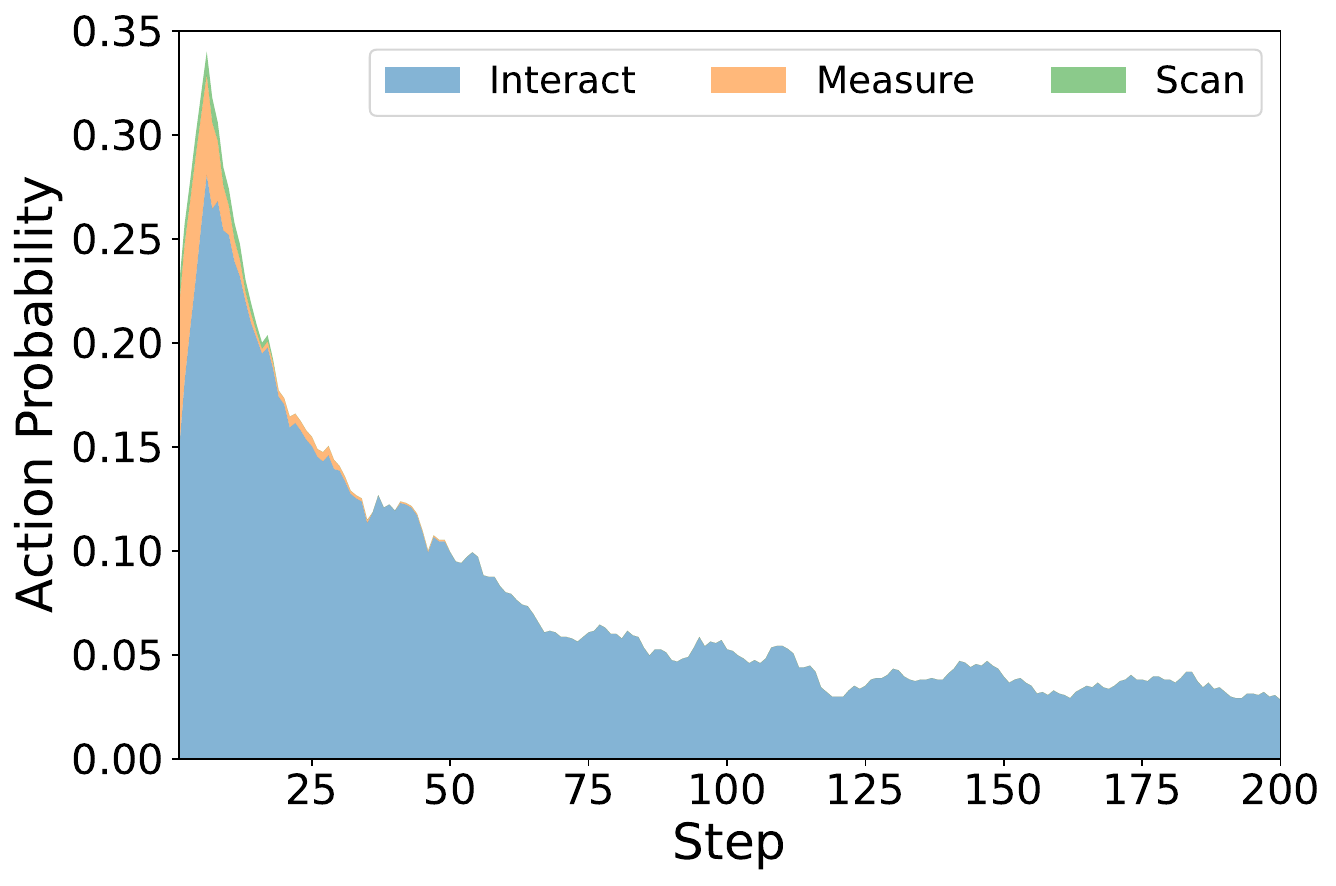}  
        \caption{Qwen3}
    \end{subfigure}
    \caption{Action-frequency profiles over the first $T{=}200$ steps under the full modifier setting. We plot only non-movement actions (\textsc{Interact}, \textsc{Measure}, \textsc{Scan}). Early spikes reflect information gathering and experimentation, while later low mass indicates trajectories dominated by movement.}
    \label{fig:action_profile}
    \postspace
\end{figure*}
\subsection{Main Results}
\label{sec:main_results}

To investigate the impact of real-world circumstances on LLM agents, we measure performance under the full set of deployment-like modifiers, where partial observability, noisy sensing, non-stationarity, and agent-state drift are simultaneously active. The results are summarized in Table~\ref{tab:main_res}. Overall, performance degrades as the grid grows from $6\times6$ to $10\times10$: all frontier models exhibit decreasing accuracy (\textsc{Qwen3} largely fails across grid sizes, underscoring the need for strong reasoning and adaptation to solve these tasks). Notably, the best performing model depends on the metric and grid size: \textsc{Gemini-3 Pro} attains the highest accuracy on $6\times6$ and $10\times10$, whereas \textsc{Gemini-3 Flash} is best on $8\times8$. This ranking instability is consistent with our setting: longer horizons increase exposure to shifts, drift, and evolving hazards, so small differences in online adaptation can dominate. This is consistent with long-horizon agent evaluations, where performance declines as the required number of steps increases \citep{wang2025odysseybench}.

A key takeaway is that \emph{accuracy alone} does not capture deployment-like behavior. Although we never explicitly instructed agents to minimize steps or maximize score 
, several models appear to partially infer these objectives and trade them off against completion. 
\textsc{GPT-5 mini} consistently has the lowest step counts across grid sizes and the best scores at $6\times6$ and $10\times10$, despite not leading in accuracy, suggesting a more efficiency- and reward-aware policy that avoids costly actions and long exploratory detours. In contrast, \textsc{GPT-5.2} remains competitive in accuracy on smaller grids but uses many more steps and its score degrades sharply with grid size, indicating that extended exploration and miscalibrated interaction under non-stationarity can accumulate large long-run costs (e.g., hazards and expensive sensing). 
The \textsc{Gemini} models show a different trade-off: \textsc{Gemini-3 Pro} attains the best $8\times8$ score with moderate steps, while \textsc{Gemini-3 Flash} achieves higher success but more negative scores, consistent with a more aggressive, penalty-prone strategy. Together, these results highlight that robust real-world readiness requires not only solving the task, but also inferring and optimizing implicit objectives under uncertainty and shifting dynamics.

\begin{figure*}[th]
    \centering
        \includegraphics[width=0.9\linewidth]{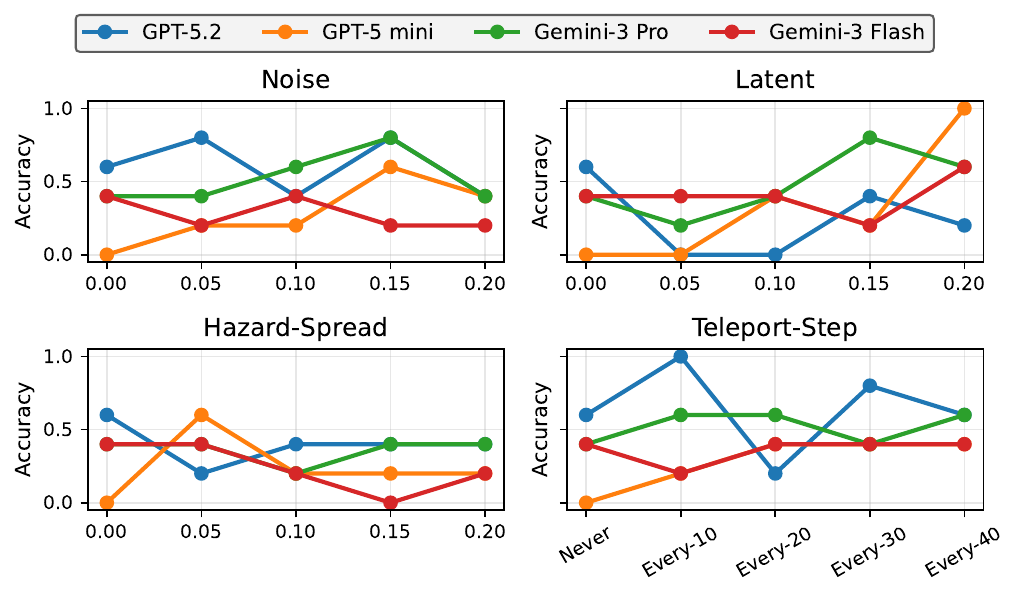}       
    \caption{Single-stressor ablations where all modifiers are disabled except the one being swept. Each panel sweeps one factor (\texttt{Noise}, \texttt{Latent}, \texttt{Hazard-Spread}, \texttt{Teleport-Step}) and reports win rate per model, averaged over 5 random episodes per data point, isolating sensitivity to individual uncertainty and non-stationarity sources.}
    \label{fig:abel}
    \postspace
\end{figure*}
\subsection{Strategic Behavior of LLM Agents}
\label{sec:st}

To compare strategic behavior beyond aggregate success and score, we analyze how LLM agents allocate actions over time within an episode. For each model, we collect trajectories from the main setting (all modifiers active) and record the action taken at every timestep. We group actions into a small set of interpretable categories (\textsc{Move} in four directions, \textsc{Scan}, \textsc{Measure}, and \textsc{Interact}) and compute an action-frequency profile over steps. 
%
Figure~\ref{fig:action_profile} visualizes these profiles for all five models. For readability, we plot only the probabilities of the three non-movement actions (\textsc{Scan}, \textsc{Measure}, \textsc{Interact}). 
Thus, higher mass in Figure~\ref{fig:action_profile} corresponds to spending more steps on explicit information acquisition or interaction, whereas lower mass indicates most steps are movement.

Several consistent patterns emerge. First, the frontier models exhibit a clear \emph{sense-then-act} signature: \textsc{Scan} and \textsc{Measure} are concentrated in the earliest steps and quickly decay toward zero, after which behavior is dominated by movement with occasional \textsc{Interact}.  
This indicates that agents generally front-load uncertainty reduction, then switch to navigation and targeted interaction once they have a partial internal map. Second, the \emph{magnitude} of early information gathering differs sharply across models. \textsc{GPT-5 mini} allocates substantially more probability mass to \textsc{Scan}/\textsc{Measure} early in the episode than the other frontier models, and later increases \textsc{Interact} as the episode progresses. This suggests an upfront sensing strategy that reduces later exploration and penalties, consistent with its low steps and higher scores in Table~\ref{tab:main_res} despite not always leading in success rate. In contrast, \textsc{GPT-5.2}, \textsc{Gemini-3 Pro}, and \textsc{Gemini-3 Flash} devote less mass to explicit sensing, rely more on movement and local observations, and keep \textsc{Interact} relatively modest, suggesting more conservative engagement with tiles.

Finally, \textsc{Qwen3} exhibits a qualitatively different regime: it spends an unusually large fraction of early steps on \textsc{Interact}, with minimal \textsc{Scan}/\textsc{Measure} (expending most of its energy early in the episode). This behavior is consistent with myopic trial-and-error interaction under underspecified rules, which is especially costly in our setting because adverse outcomes (e.g., hazards) compound over time and are exacerbated by non-stationarity and noise. This kind of strategy divergence is also observed in prior agent evaluations, where different LLMs adopt distinct behaviors when confronting recurring difficulties \citep{anupam2025browserarena}. 
Overall, these profiles highlight that success depends not just on completion but on how agents trade off movement, information gathering, and risky interaction. Despite no explicit instruction, the action profiles (notably for \textsc{GPT-5 mini}) suggest that models may recognize implicit objectives like minimizing steps and maximizing score to varying degrees, but differ in how effectively they translate that recognition into behavior. 
We also analyze how agents’ strategies shift after acquiring keys (i.e., as they become closer to the goal) in the Appendix.
Given \textsc{Qwen3}'s consistently poor performance, we omit it from the remaining analyses and focus on the other LLMs

\begin{figure*}[th!]
    \centering
        \centering
        \includegraphics[width=0.8\linewidth]{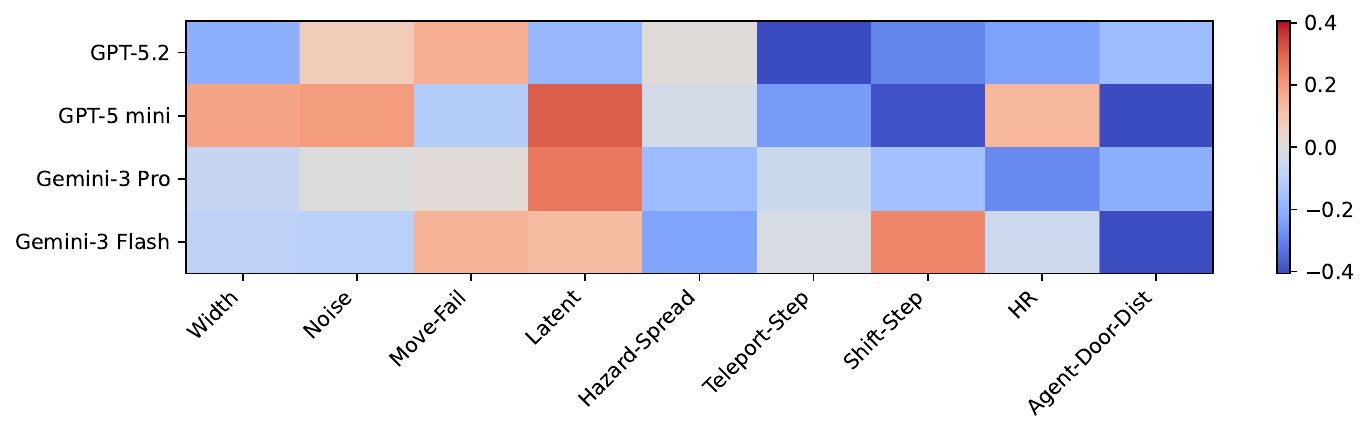}  
    \caption{Logistic regression attributions for predicting \textsc{Win} vs.\ \textsc{Lose} per LLM. Each cell is a coefficient on one of nine normalized features (positive increases win likelihood), trained per model on all episodes pooled from main results and ablations.}
    \label{fig:heat}
    \postspace
\end{figure*}
\subsection{Single-Stressor Ablation}
\label{sec:ablations}
To isolate the causal effect of each deployment stressor, we run controlled ablations that deactivate all other modifiers and sweep a single factor while measuring accuracy. Concretely, we sweep (i) observation noise (\texttt{Noise}), (ii) the fraction of latent cells (\texttt{Latent}), (iii) hazard spread probability (\texttt{Hazard-Spread}) over the same range, and (iv) the teleport schedule (\texttt{Teleport-Step}) from \emph{Never} to periodic teleports. We report performance averaged over 5 randomly generated episodes per data point (i.e., for each feature value).
Figure~\ref{fig:abel} reveals strongly non-monotonic and model-specific sensitivities. For \texttt{Noise}, moderate corruption can \emph{improve} accuracy for \textsc{GPT-5.2} and \textsc{Gemini-3 Pro} (peaking around mid-range noise) while \textsc{Gemini-3 Flash} degrades more noticeably at higher noise, suggesting different calibration of when to verify versus when to proceed. These observations are aligned with prior evidence that intermediate noise levels can act as a functional regularizer and improve adaptation under uncertainty \citep{findling2024computation}. 
For \texttt{Latent}, the models diverge even more: \textsc{GPT-5.2} drops sharply under small-to-moderate latent fractions, whereas \textsc{Gemini-3 Pro} remains comparatively robust and even improves at intermediate levels, consistent with more effective hypothesis revision and information acquisition when the map contains hidden structure. \textsc{GPT-5 mini} shows the largest variance, performing poorly at low latent fractions but improving sharply at the highest setting, suggesting it becomes effective only when uncertainty is salient enough to trigger systematic probing (consistent with its \textsc{Measure} usage increasing monotonically from $0$ to $2.6$ on average as the latent fraction rises). Overall, our results suggest that agents can benefit from hypothesis-testing actions, consistent with prior work showing that coupling decision-making with explicit information-seeking improves robustness under partial observability \citep{fang2025information}.

The dynamic-environment stressors show that ``harder'' does not always mean ``worse,'' and that disruption can sometimes act as an exploration aid. 
This is consistent with prior work showing that explicit returns/resets can improve long-horizon exploration by enabling progress from diverse states without repeatedly re-traversing the same prefixes \citep{ecoffet2021first}.
Under \texttt{Hazard-Spread}, increasing spread generally harms accuracy and can induce brittle collapse for \textsc{Gemini-3 Flash} at higher levels, while \textsc{Gemini-3 Pro} is comparatively stable across the sweep. Under \texttt{Teleport-Step}, frequent teleports can substantially \emph{help} \textsc{GPT-5.2} (near-perfect accuracy when teleporting every 10 steps) but can also hurt at intermediate frequencies, suggesting that some schedules provide useful ``free exploration'' while others repeatedly invalidate local plans. In contrast, \textsc{GPT-5 mini} improves mainly as teleports become less disruptive (better at longer intervals), reflecting a more plan-committed policy. Overall, these single-factor profiles explain why aggregate rankings in the full setting can fluctuate: success depends not only on base reasoning strength, but on whether a model’s strategy matches the specific uncertainty and non-stationarity regime—and whether it can infer the need to trade off completion with implicit objectives like efficiency and score. 

\subsection{Feature Attribution for Win/Loss Outcomes}
\label{sec:lr_attrib}
To better understand \emph{which conditions} drive success versus failure, we aggregate all episodes from the main evaluation and ablations and featurize each instance with a $9$-dimensional vector capturing normalized environment and dynamics parameters: \texttt{Width}, \texttt{Noise}, \texttt{Move-Fail}, \texttt{Latent}, \texttt{Hazard-Spread}, \texttt{Teleport-Step}, \texttt{Shift-Step}, \texttt{HR}, and \texttt{Agent-Door-Dist}. Here \texttt{HR} measures how many rule tiles have at least one hazard neighbor, capturing local ``riskiness''. For each model, we then fit a logistic regression classifier to predict \textsc{Win} vs.\ \textsc{Lose} from these features, achieving $\sim$60\% accuracy on average, and visualize the learned coefficients as a heatmap (Figure~\ref{fig:heat}).

The coefficient patterns reveal several consistent drivers of failure that align with our qualitative observations. Across models, larger \texttt{Agent-Door-Dist} tends to be negatively associated with winning, indicating that long-horizon navigation difficulty is a strong predictor of collapse even in the simplified linear model. Similarly, increased \texttt{Hazard-Spread} is generally harmful, punishing myopic exploration. In contrast, \texttt{Latent} often shows a positive association with success, suggesting that (for these agents) environments with more hidden structure can \emph{improve} outcomes when they trigger more systematic probing and hypothesis revision rather than purely reactive movement.

Importantly, the heatmap also highlights model-specific sensitivities that help explain the ranking instabilities seen in Table~\ref{tab:main_res}. \textsc{GPT-5.2} shows stronger negative associations with disruption-related features (notably \texttt{Teleport-Step} and \texttt{Shift-Step}), consistent with policies that degrade when mid-episode dynamics invalidate prior beliefs. \textsc{GPT-5 mini} places more positive weight on ``information-rich'' signals (e.g., \texttt{Latent} and \texttt{HR}), matching its tendency to front-load sensing and to behave more efficiency-aware even when its raw success rate is not the highest. The \textsc{Gemini} models appear less sensitive to some nuisance factors (e.g., \texttt{Noise} and \texttt{Move-Fail} are closer to neutral in places) but differ in how strongly they are affected by non-stationarity, which is consistent with their divergent trade-offs between completion and penalty avoidance. Overall, these results reinforce that robust deployment hinges on \emph{adapting strategy} to specific uncertainty and non-stationarity regimes rather than relying on nominal task-solving ability alone.


\section{Related Work}

\paragraph{Agentic LLM evaluation and tool robustness.}
A growing literature evaluates LLMs as multi-step tool-using agents, focusing on stateful interaction and robustness to tool failures. ToolEmu scales testing via LM-emulated tools and risk scoring \cite{ruan2023identifying}. $\tau$-bench and ToolSandbox introduce persistent state, simulated users, and trajectory-level metrics that expose brittleness \cite{yao2024tau,lu2025toolsandbox}, while Tools Fail targets silent tool errors and recovery \cite{sun2024tools}. Hell or High Water and NoisyToolBench stress replanning under external failures and ambiguous instructions \cite{wang2025hell,wang2025learning}. In contrast, our benchmark jointly instantiates partial observability, noisy sensing, non-stationarity, and agent-state drift in a single controllable environment.

\paragraph{Partial observability, latent-state inference, and shifting dynamics.}
Decision making under uncertainty (e.g., POMDPs) formalizes latent-state inference from noisy, partial observations and the role of information-gathering actions \cite{kaelbling1998planning}. Benchmarks such as MiniGrid/MiniWorld and TextWorld, ALFWorld, and ScienceWorld offer scalable long-horizon testbeds \cite{chevalier2023minigrid,cote2018textworld,shridhar2020alfworld,wang2022scienceworld}. Non-stationary RL studies adaptation to changing dynamics \cite{padakandla2020reinforcement}, and embodied benchmarks like HAZARD target dynamically changing environments \cite{zhou2024hazard}. We unify these pressures in an LLM-agent evaluation with controllable perturbations that support systematic sweeps and causal ablations.

\paragraph{Goal and preference inference under implicit objectives.}
Another thread studies inferring what is \emph{wanted} when objectives are underspecified. Cooperative IRL treats alignment as a partial-information game where the agent infers human reward while acting \cite{hadfield2016cooperative}, and language-centric work infers latent preferences from context \cite{lin2022inferring}. HandMeThat highlights ambiguity resolution in instruction following using physical/social cues \cite{wan2022handmethat}. We share this goal-inference spirit, but focus on partially implicit objectives (e.g., efficiency/score trade-offs) under unreliable, non-stationary interfaces.

\paragraph{Synthetic, controllable benchmarks for diagnosis.}
Synthetic environments remain valuable for exact ground truth and principled capability isolation. TextWorld and MiniGrid support modular/procedural instance generation \cite{cote2018textworld,chevalier2023minigrid}, and ToolEmu scales agent testing via emulated tool execution \cite{ruan2023identifying}. Following this philosophy, our compact grid game enables evaluation under combined perturbations, single-stressor sensitivity sweeps, and feature-based attributions that reveal model-specific brittleness beyond headline success rates.

\section{Conclusion}
We presented a controllable interactive benchmark that stress-tests LLM agents beyond ``clean-interface'' assumptions by combining partial observability, noisy sensing, non-stationarity, and agent-state drift in a long-horizon grid game. Across five state-of-the-art models, we observed a consistent gap between nominal task-solving and deployment-like robustness: performance generally degrades with scale, rankings are unstable across regimes, and strategy often matters more than raw capability. Despite never explicitly optimizing for efficiency or score, agents exhibited clear trade-offs between completion, steps, and penalty avoidance, suggesting partial sensitivity to implicit objectives. 

Our analyses make these gaps diagnostic. Action profiles reveal systematic differences in how models allocate effort between movement, information gathering, and risky interaction over time; single-stressor sweeps expose strongly non-monotonic and model-specific sensitivities (e.g., settings where added noise or disruption helps by altering exploration), explaining why aggregate rankings fluctuate under the full setting. Finally, simple logistic models trained on normalized stressor features predict win/loss yield interpretable coefficients, highlighting which factors most reliably drive failures for each model. Together, these results motivate future work on (i) uncertainty- and cost-aware verification policies that activate probing only when beneficial, (ii) online change detection with rapid replanning/reset under shifts and capability drift, and (iii) multi-objective training and evaluation that explicitly balances completion, efficiency, and safety costs rather than optimizing a single success signal.

\bibliography{custom}

\appendix

\begin{figure*}[th!]
    \centering
        \centering
        \includegraphics[width=0.8\linewidth]{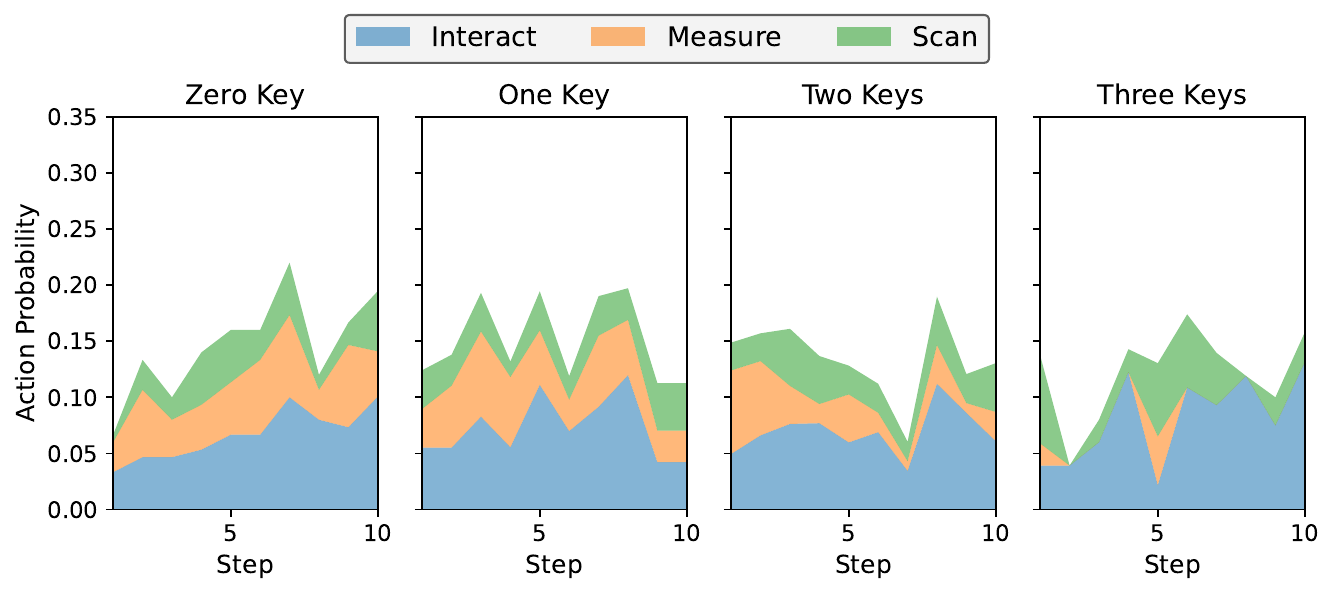}  
    \caption{Post-key action profile (10-step window), conditioned on keys collected for GPT-5.2.}
    \label{fig:gpt-5-perk}
    \postspace
\end{figure*}
\begin{figure*}[th!]
    \centering
        \centering
        \includegraphics[width=0.8\linewidth]{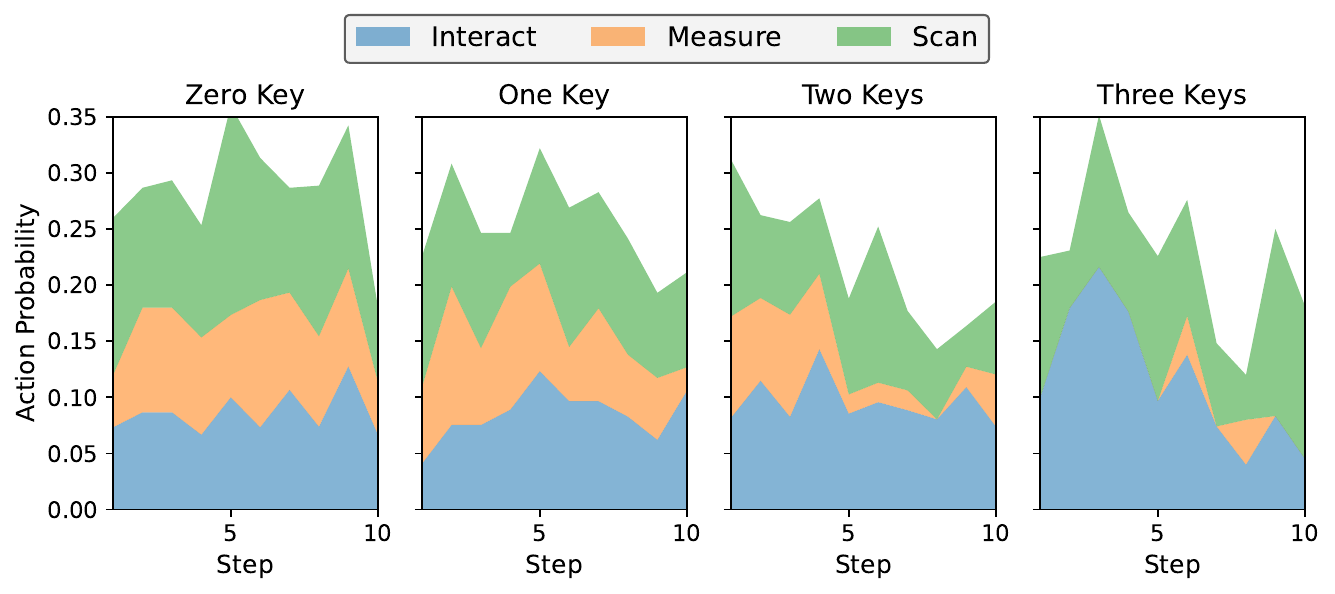}  
    \caption{Post-key action profile (10-step window), conditioned on keys collected for GPT-5 mini.}
    \label{fig:gpt-5-mini-perk}
    \postspace
\end{figure*}
\section{Prompt}
We provide the system and user prompts used for our LLM agent player in Prompts~\ref{prompt:player-system} and ~\ref{prompt:player-user}, respectively.
\begin{prompt}[title={\footnotesize\texttt{The player system prompt}}, label=prompt:player-system]
You are an embodied decision-making agent playing an interactive gridworld.\\
Your job is NOT just to finish, but to act robustly under uncertainty:\\
- The world may be underspecified and change over time.\\
- Observations may be noisy.\\
- Actions have costs.\\
\\
GAME MECHANICS / CONTROLS (read carefully):\\
- The world is a bounded grid. You see only a local window around you.\\
- Your avatar is shown as $\{\blacktriangle,\blacktriangledown,\blacktriangleleft,\blacktriangleright\}$ which also indicates the direction you are FACING.\\
- Movement actions both (a) move you one cell if not blocked AND (b) set your facing direction.\\
  * MOVE\_N sets facing north, MOVE\_S south, MOVE\_W west, MOVE\_E east.\\
- INTERACT affects the SINGLE tile directly in front of your facing direction (the adjacent cell).\\
  * If the tile in front is D (door): INTERACT attempts to open it. The door opens only if the scenario requirement is met (e.g., collect enough keys).\\
  * If the tile in front is R (rule tile): INTERACT triggers the hidden rule. The R tile may transform into outcomes like k (key), . (empty), or h (hazard). This behavior may change over time.\\
  * If the tile in front is H (hazard): INTERACT attempts to neutralize it at extra energy cost; may fail.\\
  * Otherwise INTERACT usually has no effect.\\
- SCAN temporarily increases your view radius for a step but costs energy.\\
- MEASURE (if available) collapses nearby latent tiles ($\circ$) into real tiles.\\
\\
COSTS / RISKS:\\
- Some actions consume energy (SCAN, MEASURE, INTERACT).\\
- Observations may be noisy or misleading. Also, the environment can shift.\\
\\
OUTPUT REQUIREMENT:\\
- Output EXACTLY ONE action token from the provided AVAILABLE ACTIONS list.\\
- Do not output any explanation or extra text.\\
\\
Output ONLY the action string (no extra text).
\end{prompt}
\begin{prompt}[title={\footnotesize\texttt{The player user prompt}}, label=prompt:player-user]
Collect 3 key fragments (k), then open the exit door (D) by INTERACTing with it while facing it.\\
\\
OBSERVATION (partial, local):\\
\{observation\}\\
\\
RECENT EVENTS:\\
\{recent-events\}\\
\\
PREVIOUS ACTIONS (recent):\\
\{previous-actions\}\\
\\
AVAILABLE ACTIONS:\\
\{action-space\}\\
\\
Choose exactly ONE action from AVAILABLE ACTIONS and output only that action string.
\end{prompt}

\subsection{Progress-Conditioned Strategic Shifts}
\label{sec:appendix_key_conditioned_strategy}
\begin{figure*}[th!]
    \centering
        \centering
        \includegraphics[width=0.8\linewidth]{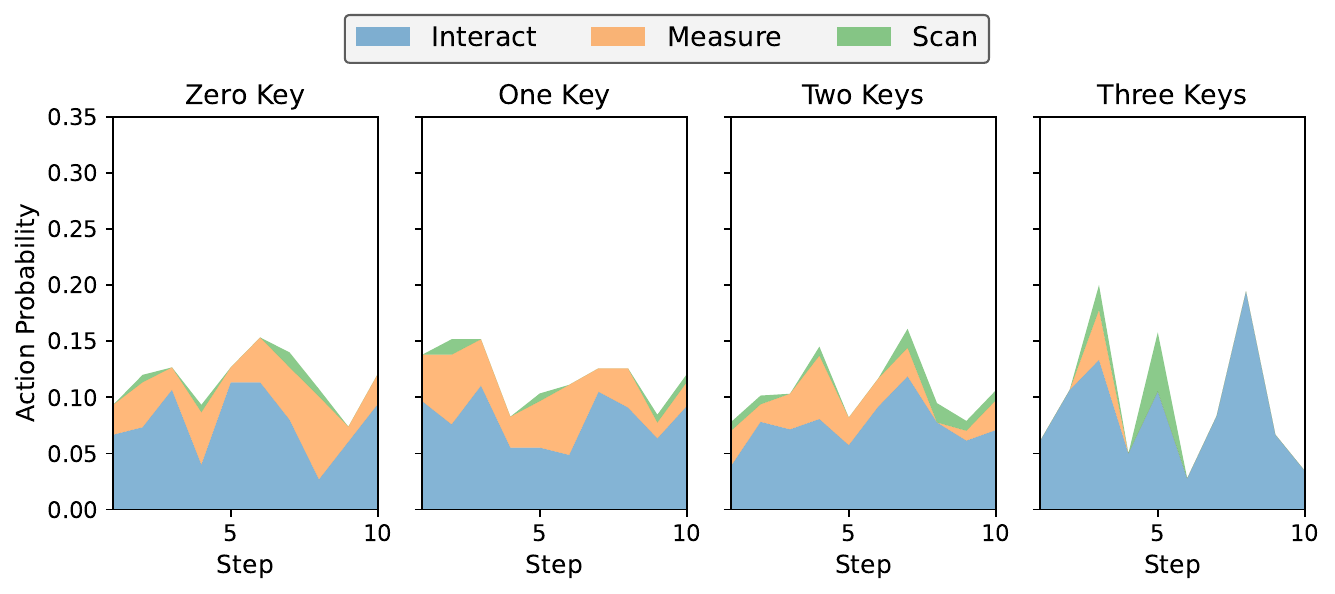}  
    \caption{Post-key action profile (10-step window), conditioned on keys collected for Gemini-3 Pro.}
    \label{fig:gemini-pro-perk}
    \postspace
\end{figure*}

\begin{figure*}[th!]
    \centering
        \centering
        \includegraphics[width=0.8\linewidth]{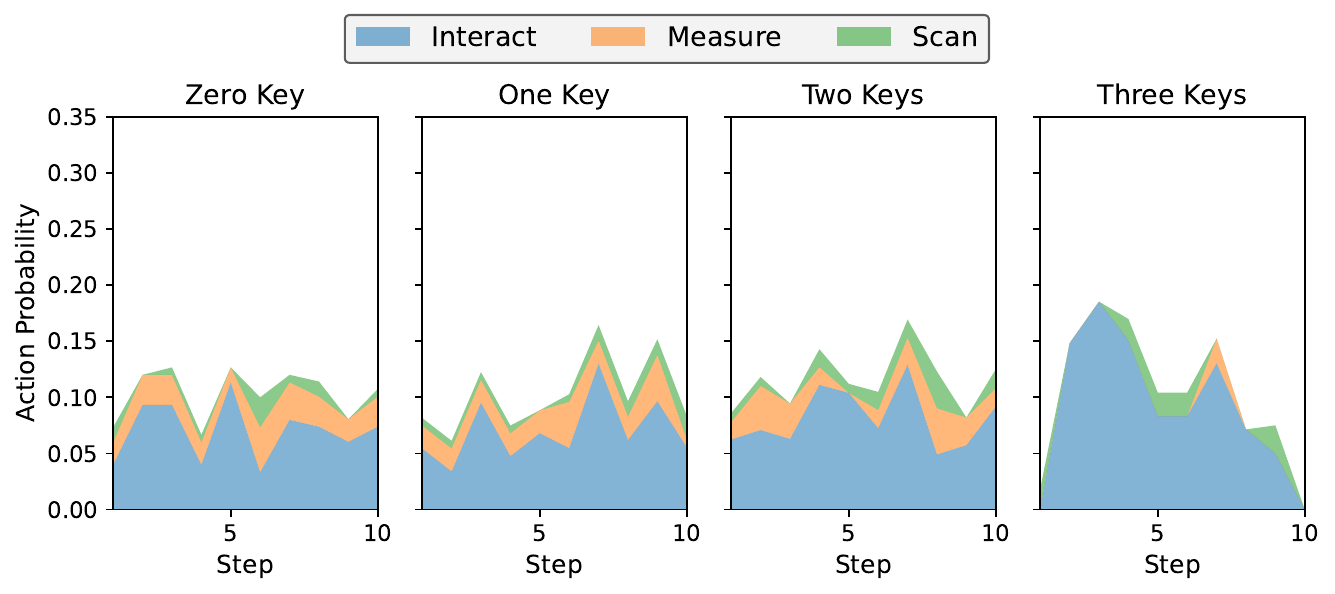}  
    \caption{Post-key action profile (10-step window), conditioned on keys collected for Gemini-3 Flash.}
    \label{fig:gemini-flash-perk}
    \postspace
\end{figure*}

\begin{figure*}[th!]
    \centering
        \centering
        \includegraphics[width=0.8\linewidth]{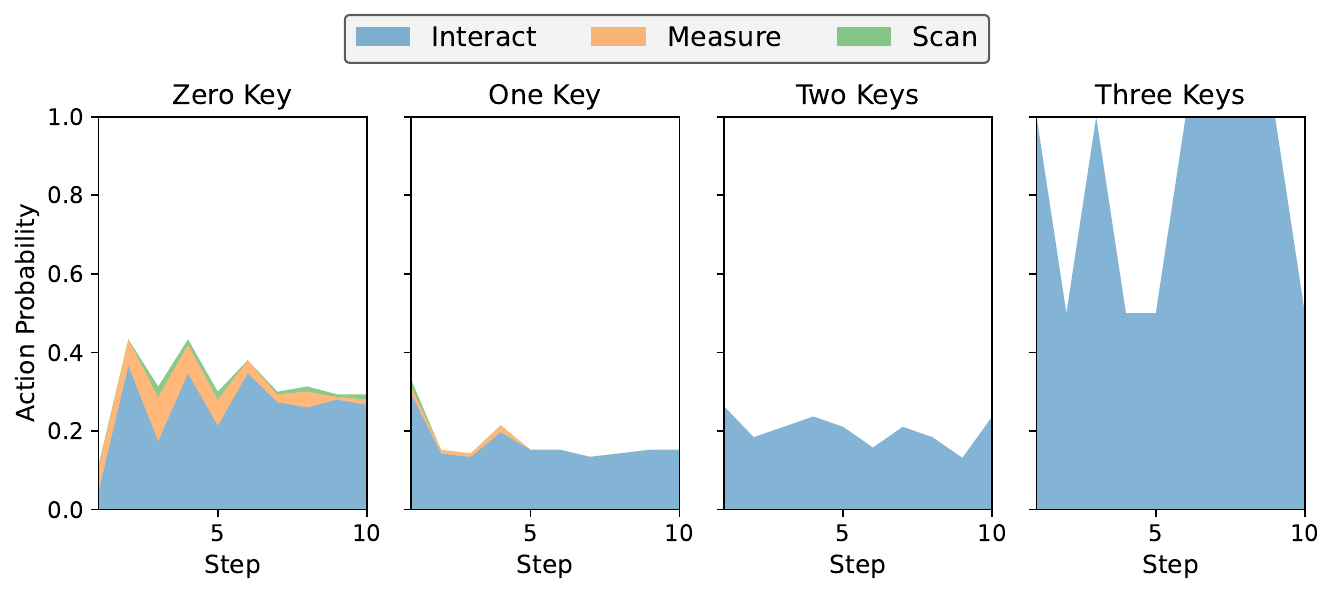}  
    \caption{Post-key action profile (10-step window), conditioned on keys collected for Qwen3.}
    \label{fig:qwen-perk}
    \postspace
\end{figure*}

Using the same trajectories from Section \ref{sec:st}, we further analyze how agents change their behavior as they make progress toward the goal. Concretely, for each episode we identify the first timestep at which the agent’s key count increases to a given level ($0,1,2,3$). We then align episodes at that milestone and compute an action-frequency profile over the subsequent 10 steps. As in Section \ref{sec:st}, we aggregate actions into \textsc{Move}, \textsc{Scan}, \textsc{Measure}, and \textsc{Interact}, and plot only the three non-movement probabilities for readability. Figures \ref{fig:gpt-5-perk}, \ref{fig:gpt-5-mini-perk}, \ref{fig:gemini-pro-perk}, \ref{fig:gemini-flash-perk}, and \ref{fig:qwen-perk} report these progress-conditioned profiles for the five LLM agents.

Across frontier models, the profiles reveal a consistent stage structure. When agents have \emph{zero keys}, \textsc{Scan}/\textsc{Measure} usage is comparatively higher, reflecting early uncertainty reduction and map acquisition. As key count increases, explicit sensing generally declines and behavior shifts toward navigation with occasional \textsc{Interact}, indicating a transition from exploration to goal-directed execution. This transition is most pronounced near \emph{three keys}, where several models sharply increase \textsc{Interact} immediately after acquiring the final key, consistent with actively seeking and engaging the door \(D\) to terminate the episode. Notably, \textsc{GPT-5 mini} exhibits the strongest early sensing mass and retains nontrivial probing even after collecting keys, consistent with an ``invest upfront, execute efficiently'' strategy that aligns with its low step counts and competitive scores in Table~\ref{tab:main_res}. In contrast, \textsc{GPT-5.2} and the \textsc{Gemini} models devote less probability mass to explicit sensing throughout, relying more heavily on movement and local observations, and primarily spike \textsc{Interact} in the late stage.

\textsc{Qwen3} again exhibits a qualitatively different pattern: it allocates unusually high probability to \textsc{Interact} across all progress levels, and becomes almost entirely dominated by \textsc{Interact} after reaching three keys. This is consistent with myopic, trial-and-error engagement with nearby tiles rather than a progress-aware shift toward navigation and task closure, and helps explain its poor performance under deployment-like stressors where adverse outcomes compound over time. Overall, these progress-conditioned profiles show that robust performance depends not only on which actions agents choose in aggregate, but on whether they adapt those choices as the episode transitions from exploration (uncertainty reduction) to exploitation (safe, efficient completion).


\end{document}